\documentclass{IOS-Book-Article}
\usepackage{mathptmx}
\usepackage{soul}\setuldepth{article}

\usepackage{url}
\usepackage{amsmath}
\usepackage{graphicx}
\usepackage[T1]{fontenc}
\usepackage{fixltx2e}
\usepackage{subcaption}
\usepackage[table]{xcolor}
\usepackage{url}
\usepackage{physics}
\usepackage{tikz-cd}
\usepackage{hyperref}
\usepackage[square,sort,comma,numbers]{natbib}

\newcommand{\secref}[1]{Section~\ref{sec:#1}}

\newcommand{\figref}[1]{Figure~\ref{fig:#1}}
\newcommand{\figstworef}[2]{Figures~\ref{fig:#1} and~\ref{fig:#2}}
\newcommand{\tabref}[1]{Table~\ref{tab:#1}}

\DeclareRobustCommand{\DE}[3]{#2}

\newcommand{\F}{F$_{01}$}
\newcommand{\Fw}{F$_{\mu}$}

\definecolor{codegreen}{rgb}{0,0.6,0}
\definecolor{codegray}{rgb}{0.5,0.5,0.5}
\definecolor{codepurple}{rgb}{0.58,0,0.82}
\definecolor{backcolour}{rgb}{0.95,0.95,0.92}

\usepackage[framemethod=TikZ]{mdframed}
\usepackage{xcolor}
\usepackage{listings}
\usepackage{caption}
\newmdenv[%
    innertopmargin=-10pt,
    innerbottommargin=-10pt,
    innerleftmargin=3pt,
    innerrightmargin=-2pt,
    skipabove=0pt,
    skipbelow=0pt,
    backgroundcolor=blue!5,
    linecolor=black,
    outerlinewidth=0.01pt,
    font=\footnotesize,
]{promptbox}
\makeatletter
\let\orig@promptbox=\promptbox
\def\promptbox{
  \@ifnextchar[{\promptbox@opt}{\orig@promptbox}
}
\def\promptbox@opt[#1]{
  \orig@promptbox[frametitle={#1}]
}
\makeatother

\begin{document}

\begin{frontmatter}

\title{``I'd Like to Have an Argument, Please'': Argumentative Reasoning in Large Language Models}

\author[A,B]{\fnms{Adrian} \snm{de Wynter}\orcid{0000-0003-2679-7241}%
\thanks{Corresponding Author: Adrian de Wynter, adewynter@microsoft.com. The final version of this paper is accepted to COMMA '24}}
and
\author[B]{\fnms{Tangming} \snm{Yuan}\orcid{0000-0003-1697-7003}}

\runningauthor{A. de Wynter and T. Yuan}
\address[A]{Microsoft}
\address[B]{The University of York}

\begin{abstract}
We evaluate two large language models (LLMs) ability to perform argumentative reasoning. 
We experiment with argument mining (AM) and argument pair extraction (APE), and evaluate the LLMs' ability to recognize arguments under progressively more abstract input and output (I/O) representations (e.g., arbitrary label sets, graphs, etc.). 
Unlike the well-known evaluation of prompt phrasings, abstraction evaluation retains the prompt's phrasing but tests reasoning capabilities. 
We find that scoring-wise the LLMs match or surpass the SOTA in AM and APE, and under certain I/O abstractions LLMs perform well, even beating chain-of-thought--we call this \emph{symbolic prompting}. 
However, statistical analysis on the LLMs outputs when subject to small, yet still human-readable, alterations in the I/O representations (e.g., asking for BIO tags as opposed to line numbers) showed that the models are not performing reasoning. 
This suggests that LLM applications to some tasks, such as data labelling and paper reviewing, must be done with care. 

\end{abstract}

\end{frontmatter}

\section{Introduction}\label{sec:intro}

Large language models (LLMs) such as GPT-4 \citep{GPT4} have shown to have spectacular accuracy on a variety of tasks. 
Hence, attempts have been made to automate more complex tasks reliant on argumentative reasoning, such as data labelling \citep{cheng2023gpt4} and scientific paper reviews \citep{liu2023reviewergpt}. 
Argumentative reasoning encompasses formal and informal logic, and requires a deep understanding of, and reasoning over, the pragmatic context. 
Hence, to understand the reliability of LLMs in these tasks we must also evaluate their argumentative reasoning capabilities. 
This goes beyond determining whether the model can generate relevant responses, and asks if it can robustly reason over the context and solve the task. 

We evaluate the argumentative reasoning capabilities of two LLMs, GPT-3 \citep{GPT3} and GPT-4 in two tasks, argument mining (AM) and argument pair extraction (APE \cite{cheng-etal-2020-ape}).\footnote{Prompts, code, and outputs are in \url{https://github.com/adewynter/argumentation-llms}} 
We do this by an \emph{abstraction evaluation}: measuring progressively more abstract input and output (I/O) representations. 
Unlike prompt-phrasing evaluations, to which LLMs are known to be sensitive \cite{lu-etal-2022-fantastically,webson-pavlick-2022-prompt}, our evaluation maintains the task description untouched, and only alters the signature of the data. 
For example, adding line numbers to the input and requesting the output to be from a specific set (e.g., $\{0, 1\}$) is a conceptually minor I/O representation change. 
This retains the task description, but requires some level of reasoning to return a correct and parseable response given the specified signatures. 
In this paper we then measure the LLMs' argumentative reasoning capabilities indirectly, by testing their ability to \emph{robustly} recognize arguments when altering the I/O.

\subsection{Findings}

In terms of raw scoring, we find that GPT-4 is able to reach SOTA performance in APE, and near-SOTA in AM. However, our analysis shows that:

\begin{enumerate}
    \item LLM scoring varies dramatically with the abstraction level, which suggests a lack of comprehension of the task. Note that across our experiments the task description remained fixed, and we only altered the I/O representation.
    \item \emph{Symbolic prompting} (that is, low-abstraction, hints-enabled inputs) bests other approaches, including chain-of-thought (CoT) \cite{ChainOfThought}. 
    \item CoT approaches are robust to abstraction, but its output distributions over abstractions are similar, which we attribute to its templatized nature.
    \item LLM scores worsen with more exemplars, indicating poor inductive reasoning capabilities. 
\end{enumerate}

We conclude that the LLMs are unable to reason reliably in an argumentative setting.

\section{Related Work}\label{sec:relatedwork}

We discuss evaluations of LLMs as it pertains to argumentative reasoning. 
For non-LLM-based approaches see \cite{lawrenceandreed}; and for a survey on reasoning in LLMs see \cite{huang-chang-2023-towards}. 
LLMs are relatively new, though they typically outperform non-LLM approaches, for example in AM \cite{10194998} and argument evaluation \cite{van-der-meer-etal-2022-will,holtermann-etal-2022-fair}, in terms of raw scoring. 
Indeed, LLMs have been observed to generalize to (read: score well in) unseen tasks without training, thus raising the question as to whether they can reason about the prompts; or are just regurgitating their training data or returning semantically-close responses. 
To some, this generalization is an indication of emergent reasoning capabilities \cite{wei2022emergent,suzgun-etal-2023-challenging}; but it has been posed that with better statistics this evidence of emergence disappears \cite{schaeffer2023are}. 
Some tests, such as GPT-4's own technical report \cite{GPT4}, tout remarkable reasoning capabilities. There is also evidence to the contrary, e.g. in code generation \cite{liu2023is}, scientific questions \cite{ChatGPTscienceengg}, first-order logic under fictional worlds \cite{saparov2023language}, and arithmetic \cite{Dziri2023FaithAF}; more generally, tasks with significant reasoning depths cause LLMs to fail \cite{anil2022exploring,valmeekam2022large}. 
It has hence been suggested that LLMs do not actually reason, but rely on heuristics (e.g., semantic similarity) \cite{patel-etal-2021-nlp}. 
Remark that these studies are limited to the prompts and versions of the models available then. 
It was suggested that LLMs are not meant for formal reasoning, and it is better to evaluate them in real-world (informal, inductive) scenarios \cite{valmeekam2022large}; yet GPT-3 cannot mimic human-like inductive reasoning \cite{humanlike}, or understand the prompts \cite{webson-pavlick-2022-prompt}. 
LLMs have also been found to not be competent in legal reasoning, due to their inability to make good arguments \cite{nguyenlaw}. 
LLMs may also retrieve dialogue acts, in line with their success as chatbots, but do not understand offers in negotiations (i.e., pragmatics) \cite{lin_toward_2023}.

\section{Methodology}\label{sec:methodology}

\subsection{Data}\label{sec:data}

\begin{figure}
\centering
\includegraphics[width = \columnwidth]{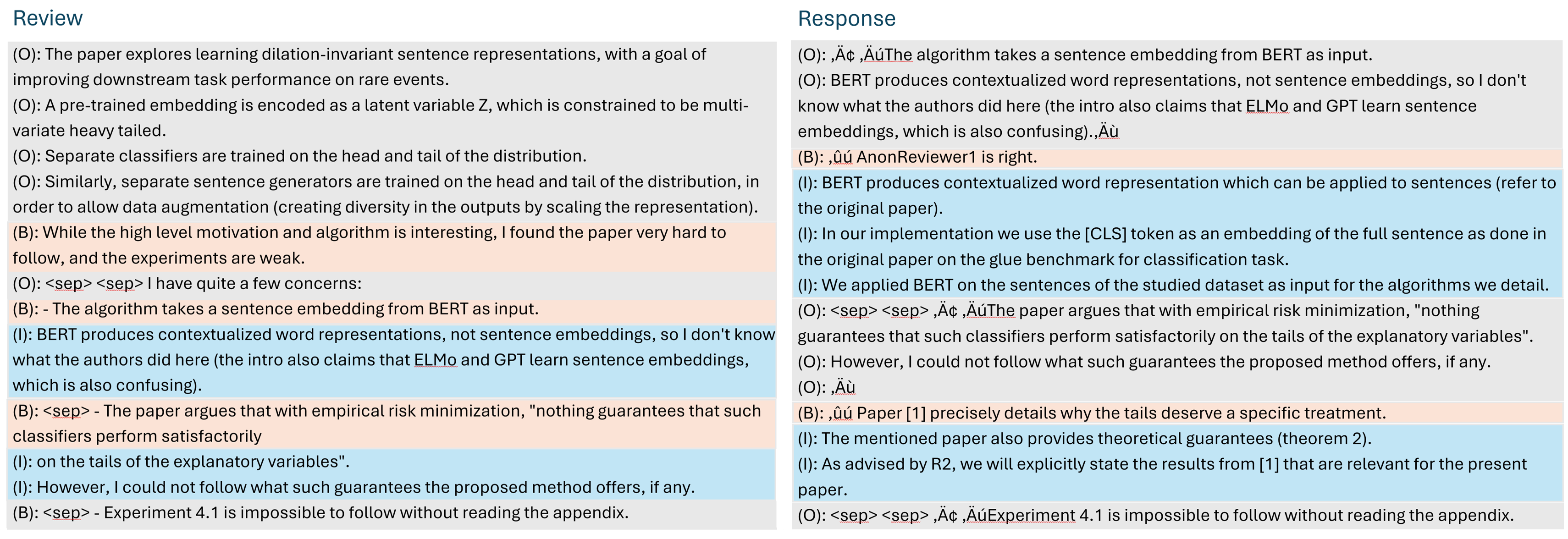}
\caption{Examples of the entries in RRv2. 
The BIO tag is in parenthesis. Highlighted in orange and blue are the ``B'' and ``I'' labels corresponding to an argument. 
In AM, review and rebuttal (response) passages are independent, and the task is to assign BIO tags to every line. 
In APE, the task is to match arguments from the review to their corresponding responses. In the above the first argument is unmatched, and the second (``BERT produces...'') pairs to the first argument from the response. 
}
\label{fig:amexample}
\end{figure}

Throughout this paper, we utilize the Review-Rebuttal Submission-v2 (RRv2) dataset \cite{cheng-etal-2020-ape}. 
It is a comprehensive corpus focused on long-distance relationships between statements, and includes both AM and APE. 
It has $4,764$ ($474$ for test) pairs of review and rebuttal passages related to scientific article submissions. 
Each passage is sentence-separated, and includes multiple arguments. 
It is human-labelled. 
For AM, each sentence is labelled with a BIO tag,\footnote{In the RRv2 corpus, the BIO tags correspond to the \underline{B}eginning, \underline{I}nner, and \underline{O}uter parts of an argument.} and the model must retrieve (label) each sentence from the review and rebuttal entries. 
In AM the distinction between review and rebuttal is irrelevant: each entry is treated as a separate point in the corpus. 
For APE, the task is to align the arguments within each review-rebuttal pair: 
every argument made by a reviewer must be mapped, when applicable, to a response from the rebuttal. 
This is normally represented as a binary matrix with overlaps \cite{bao-etal-2022-arguments,cheng-etal-2021-argument}. 
Prior to use we clean the text from tag and sentence delimiters. 
See \figref{amexample} for a sample of the corpus. 

\subsection{LLMs Evaluated}
We evaluate GPT-4 and the \textsc{text-davinci-003} variant of GPT-3 (``GPT-3.51''). 
Both models are autoregressive language models, instruction-pretrained \cite{wei2022finetuned,NEURIPS2022_b1efde53} and tuned with reinforcement learning with human feedback \cite{NEURIPS2022_b1efde53,NIPS2017_d5e2c0ad}. 
For GPT-4, there are no details released around the architecture, model size, or training data. 
It is considered better than GPT-3 at more complex tasks \cite{GPT4}. 
The variant of GPT-4 we used (``GPT-4-0613'') has a context length of $32,768$ tokens; 
GPT-3 has $4,097$ tokens.\footnote{\url{https://platform.openai.com/docs/models}}. 
Throughout our experiments, for both LLMs we set the temperature to $0.8$, the maximum return tokens based on the task, and left everything else as default. 
To account for randomness, we report the average of five calls per point to the Azure OpenAI API. %

\subsection{Prompting}

Prior to starting the work we tuned the prompt phrasing for best performance. 
When using $n$ exemplars, we used the first $n$ points from the development set. 
For AM we prompted GPT-4 with and without CoT. 
CoT conditions the LLM to work step-by-step by following a templatized process (e.g., ``Let's think about this step-by-step...''). 
It is known to provide good results in multiple reasoning tasks \cite{ChainOfThought,NEURIPS2022_8bb0d291,suzgun-etal-2023-challenging}. 
All our prompts followed the structure from GPT-4's technical report \cite{GPT4}, as we observed it produced more reliable outputs. 
For CoT we followed a tuned template (``Let's read line-by-line and solve this step by step'') indicating which sentence was being read, as well as the rationale. 
For example, ``We now read \{SENTENCE\}. It follows the previous argument, and hence it is labelled with an `I'.'' 
Sample prompts can be found in \figstworef{sampleamconcrete}{sampleamsymbolic}. 

RRv2 is measured with binary F$_{1}$ (\F{}) for APE, and micro-F$_1$ (\Fw{}) for AM \cite{cheng-etal-2020-ape}. 
Our prompts have specified a return format to signal the beginning of parsing. %

\subsection{Baselines}

Our baselines are the MLMC \cite{cheng-etal-2021-argument} and MRC-APE \cite{bao-etal-2022-arguments} models. 
MLMC approaches APE as a table-filling problem: passages are related by their pairing on a table and it relies on an especially designed encoding scheme and loss. 
MRC-APE phrases it as a reading comprehension task: first, the model does AM, and then pairs the detected arguments. 
This approach is effective when using longer-context layers, of up to $4,096$ tokens. 
We additionally consider random guessers for AM (around 33\% \Fw{}) and APE (14\% \F{}). 

\begin{figure}
\begin{subfigure}[t]{0.48\textwidth}
\centering
\begin{promptbox}
\begin{lstlisting}
Extract from the passage all the arguments.
The output should be in the form:
|begin response|
|START| line from argument 1
line from argument 1
|START| line from argument 2
line from argument 2
...
etc
|end response|
For example, 
{EXEMPLARS GO HERE}
Passage:
{PASSAGE GOES HERE}
Response:
|begin response|\end{lstlisting}
\end{promptbox}
\caption{\footnotesize{Sample concrete AM prompt. 
The model must mark every new argument with a special token (``\texttt{|START|}'') for identification. 
In APE we ask for the pairing (e.g., ``return all arguments from the response that match those of the review'').
}}
\label{fig:sampleamconcrete}
\end{subfigure}\quad\begin{subfigure}[t]{.48\textwidth}
\centering
\begin{promptbox}
\begin{lstlisting}
Extract from the passage all the arguments.
Label the beginning of every argument with a
"B". Label the rest of the argument with an
"I". Label every line that is not part of an
argument with an "O". 
The output should be in the form:
|begin response|
B
I
...
etc
|end response|
For example, 
{EXEMPLARS GO HERE}
Passage:
{PASSAGE GOES HERE}
Response:
|begin response|
\end{lstlisting}
\end{promptbox}
\caption{\footnotesize{Sample symbolic AM prompt. 
The model must return BIO tags. 
In other representations (e.g., indices) the model must to mark the B-label in parenthesis (e.g, ``(15) 16 17''). 
In APE this output is of the form ``argument lines: response lines'', and we convert into a binary matrix for scoring. 
}}
\label{fig:sampleamsymbolic}
\end{subfigure}
\caption{\footnotesize{Sample prompts for our concrete (left) and symbolic (right) settings. 
Exemplars, if any, are included in the prompt. 
For zero-shot we only specify the output representation. Note how the actual task definition and prompt structure remains unchanged, and we only alter the I/O representations.}}
\end{figure}

\subsection{Settings}

We have named our settings (i.e., representations) as \emph{concrete} and \emph{symbolic}, to distinguish the approach taken towards representing the task. 
Concrete returns full sentences, while symbolic encompasses a variety of I/O symbols. 
This is only for practical purposes: symbolic approaches cover multiple I/O representations, some of which may be easier than concrete; 
and, strictly speaking, the concrete setting is a type of symbolic representation \citep{Levesque}. 
See \tabref{representations} for a full description of the settings tested. 

The \emph{concrete} setting we instruct the LLM to return lines in text based on the prompt: in AM, it must be part of an argument, in APE, an argument pair. 
To distinguish the ``B'' and ``I'' labels, we enforce a specific return format to work with our parsing code via a special token (\texttt{|START|}). 
For scoring concrete settings we expect an exact text match. 

In \emph{symbolic} settings the LLM must return symbols (labels) based on the prompt. 
This requires more reasoning steps than in concrete settings: the LLM is solving AM \emph{and} labelling the span with an arbitrary label set defined in the prompt. 
In AM symbolic we evaluated two types of labels: BIO tags and line indices. 
For APE we evaluated line indices and the full binary matrix representation. 
We also used abstract meaning representation (AMR \cite{banarescu-etal-2013-abstract}) graphs, which are used in argument interpretation \cite{opitz-etal-2021-explainable}. %

\begin{center}
\begin{table}[h]
\centering
\begin{tabular}{|c | c | c | c |} \hline
Task & Input & Output (label set) & Abstraction  \\ \hline\hline
AM* & Text & Text and \texttt{|START|} & Lowest\\ \hline
AM* & Text with indices & Indices & Low \\ \hline
AM* & Text & Indices & Medium \\ \hline
AM* & Text & BIO tags & Medium-high \\ \hline
AM & Text with AMR graph & BIO tags & High \\ \hline
AM & AMR graph & BIO tags & Highest \\ \hline %
\end{tabular}
\caption{Input representations tested, in roughly increasing order of abstraction. 
Tasks marked with an asterisk (*) were tested with and without CoT. 
The first row is our concrete reasoning setting. 
Our ranking of abstraction is arbitrary: 
we consider the text with indices marked inline less abstract than a text without them, since the former provides a ``hint'' of what the label set is supposed to be like. 
Output representations with BIO tags as the output are more abstract, since it requires rule matching to determine the labeling. 
}\label{tab:representations}
\end{table}
\end{center}

\section{Experiments and Results}\label{sec:experiments}

We report our results comparing raw scores with respect to our settings (\secref{results}); number of exemplars (\secref{exemplars}); and I/O representations (\secref{inputrep}). 

\subsection{AM/APE: Symbolic and Concrete Reasoning}\label{sec:results}

Results for the best-performing prompts and settings are in \tabref{sample}, and a description of every setting in \tabref{representations}. 
In AM the LLMs did well but did not beat the SOTA. 
The best-performing symbolic setting had line indices included in the input representation, and requested the indices of each argument as the output representation. 
To convert to BIO tags, we instructed the model to return the ``B'' labels as indices enclosed in parentheses. 
Not including the indices in the input did not lead to an equivalently good performance. 

For APE, GPT-4 consistently bested the best-performing models ($+14\%$ \F). 
Both symbolic and concrete approaches did well with respect to the existing non-LLM-based baselines. 
We tested other approaches, such as first extracting the arguments and then matching them, but it did not yield sufficiently good results. 
Requesting a binary matrix output led to extremely poor performance ($9.88\%$ \F, below random). 
Due to token-length and budget limitations, we were unable to test CoT and AMR in APE. 

CoT approaches had generally better performance than their non-CoT counterparts, even though they use fewer exemplars. 
The only exception to this was symbolic prompting (indices inline and indices as output) where the difference was $3\%$ points. %

\begin{center}
\begin{table}[h]
\centering
\begin{tabular}{|c || c | c |} \hline
Model & AM \Fw & APE \F   \\ \hline
GPT-3 (concrete) & $39.86 \pm 0.51$ & $18.58 \pm 0.70$ \\ \hline
GPT-4 (concrete) & $64.51 \pm 0.53$ & $53.84 \pm 0.73$ \\ \hline \hline
GPT-3 (symbolic) & $62.00\pm 0.32$ & $20.15 \pm 0.91$  \\ \hline
GPT-4 (symbolic) & $70.63\pm 0.21$ & $49.85 \pm 0.96$ \\ \hline \hline
MRC-APE & $72.43$ & $39.92$ \\ \hline
MLMC    & $71.35$ & $32.81$  \\ \hline%
\end{tabular}
\caption{Results for the AM and APE tasks in our settings. 
The best-performing symbolic prompt had indices inline and indices as the output. 
We also report MLMC and MRC-APE, the two best-performing, non-LLM-based approaches for RRv2. 
GPT-4 almost matched the existing baselines in AM and bested them in APE. 
}\label{tab:sample}
\end{table}
\end{center}

\subsection{Performance and Number of Exemplars}\label{sec:exemplars}

\begin{figure}
\begin{subfigure}[t]{.48\textwidth}
\centering
\includegraphics[width = \linewidth]{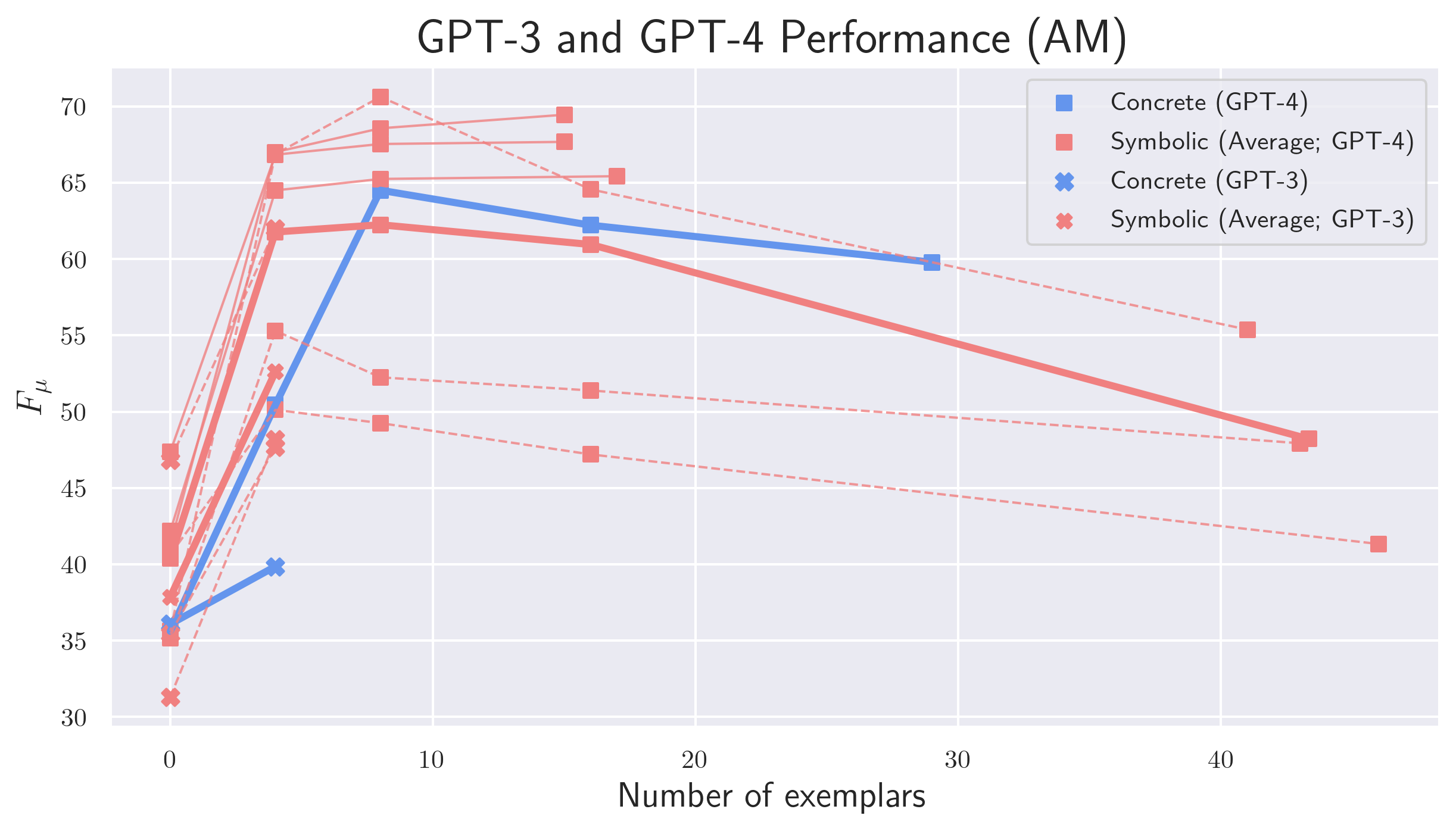}
\caption{\footnotesize{Number of exemplars versus \Fw{} in AM. 
CoT (thin solid red lines) outperformed concrete and all but one of the symbolic approaches.}}
\label{fig:shotsam}    
\end{subfigure}\quad
\begin{subfigure}[t]{.48\textwidth}
\centering    
\includegraphics[width = \linewidth]{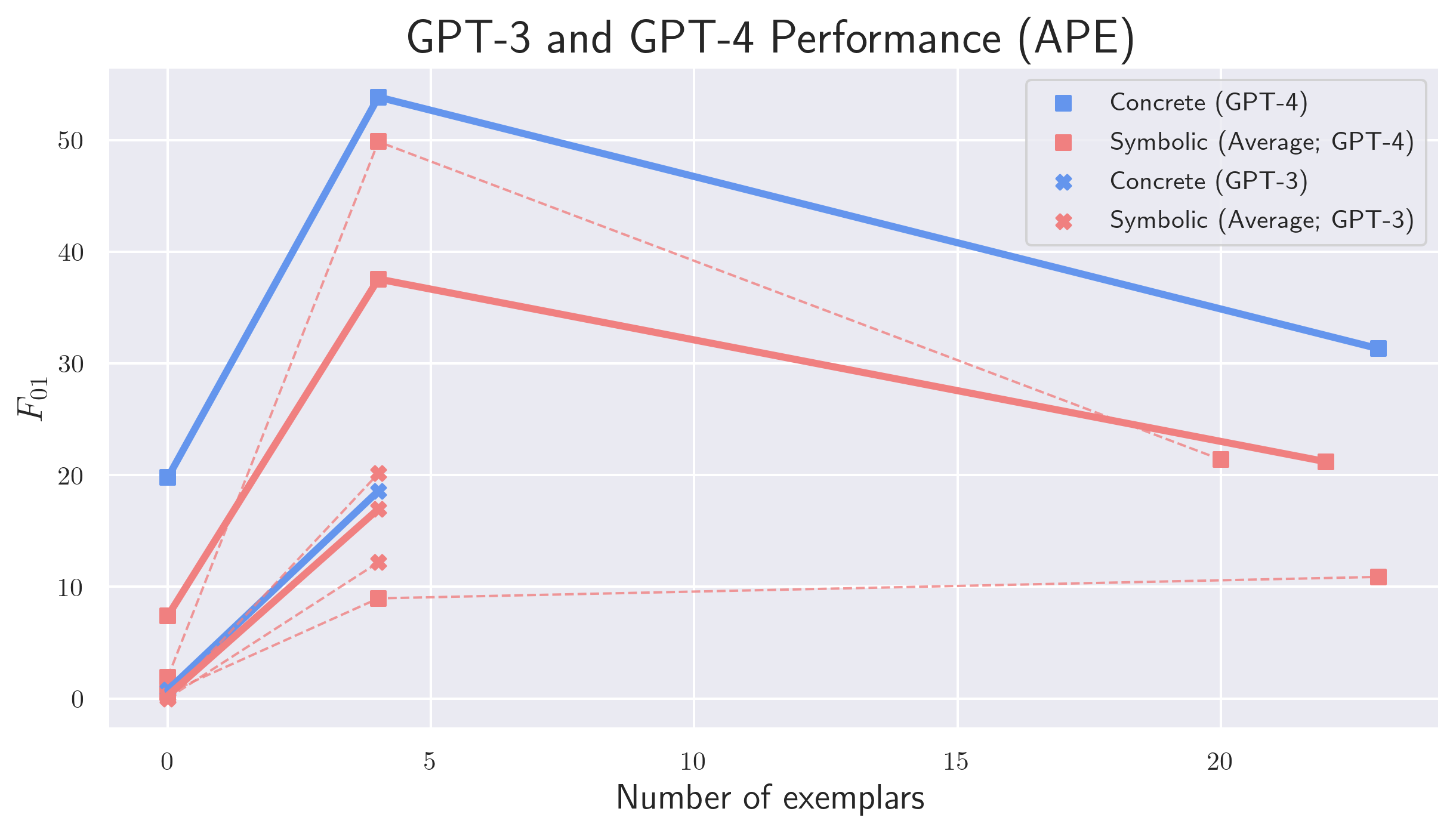}
\caption{\footnotesize{Number of exemplars versus \F{} in APE. 
We could not test CoT and AMR due to limitations on token length and budget.}}
\label{fig:shotsape}    
\end{subfigure}
\caption{Exemplar number versus score for AM and APE. In blue is the concrete approach; thick red line is the averaged performance (symbolic; dashed red lines). 
Last entry in all plots is the average maximum number of exemplars supported. 
LLM scoring in non-CoT peaked at around $4$ exemplars, and decreased afterwards. }
\end{figure}

We compared the number of exemplars ($\{0, 4, 8, 16, \tau\}$, where $\tau$ is the maximum number) with the LLM performance. 
For GPT-3, $\tau$ tended to be around $4$; for GPT-4 it varied, with an average of $44$ for symbolic, non-CoT approaches in AM ($15$ CoT, $9$ AMR) and $29$ concrete; and $22$ for both approaches in APE. 
Results are in \figstworef{shotsam}{shotsape}. 
The LLMs peaked in performance at $4$ exemplars, and their scores decreased from there. 
This was independent of the task and setting. %
We did not observe this trend in CoT. 

\subsection{Performance and Input Representation}\label{sec:inputrep}
\begin{figure}
    \centering
    \includegraphics[width=0.75\columnwidth]{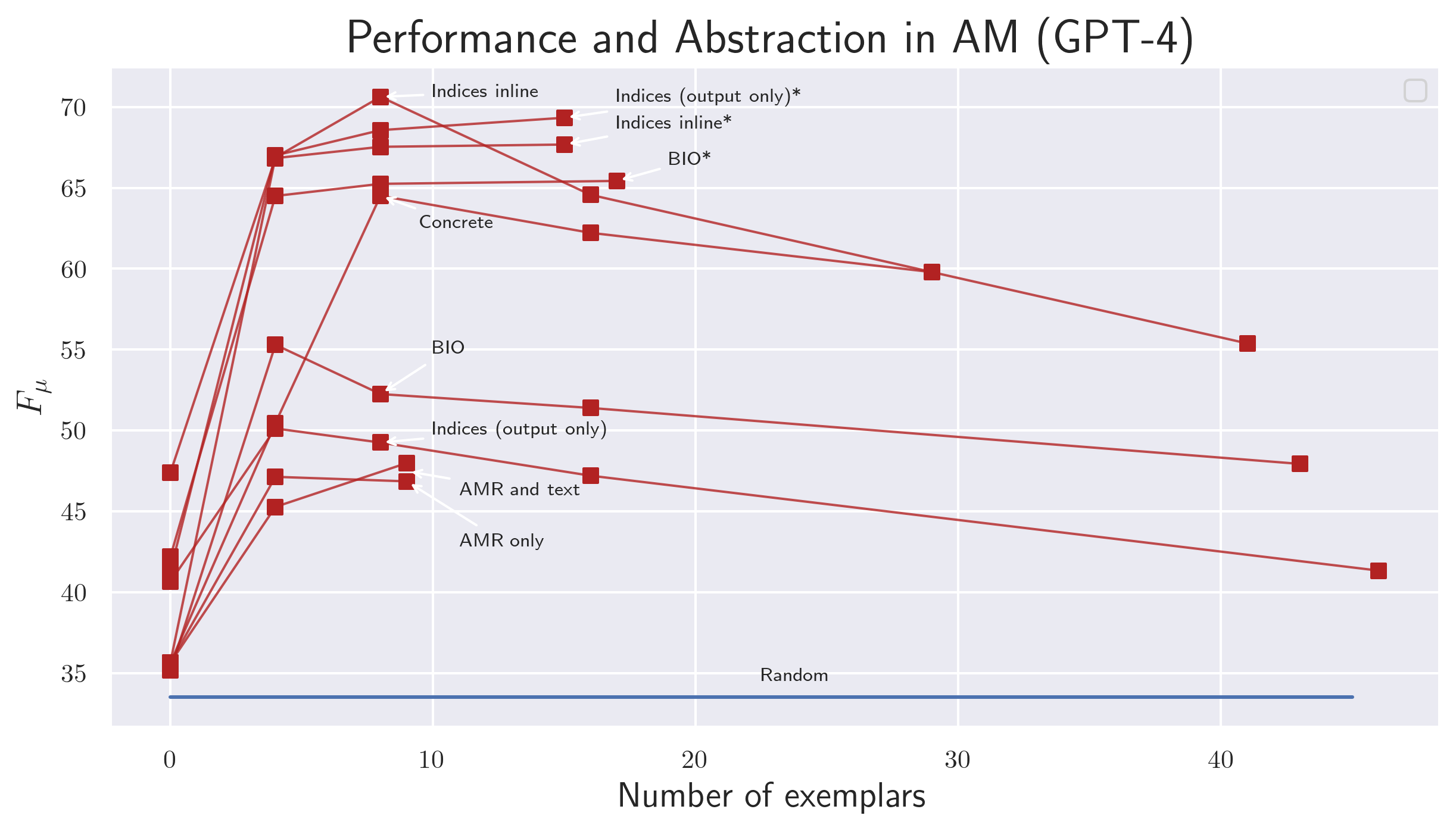}
    \caption{Effect of abstraction in the I/O representation with respect to \Fw{} in AM. 
    For non-CoT, the more abstract the input, the more difficult it is for the model to solve the task. 
    Even the most abstract representations (AMR graphs) are noticeably better than a random. 
    In CoT (top, marked with an asterisk) consistently outperformed non-CoT, even when the maximum number of exemplars supported is much lower, and regardless of abstraction level. We did not test CoT with AMR in AM due to token limitations.}
    \label{fig:abstraction}
\end{figure}

In this section we focused on GPT-4 and AM and the following representations:
text with indices inline, plain text, and with and without an AMR graph. 
There is no rigorous way to quantify the level of abstraction for these. 
However, we consider the concrete approach to be least abstract; 
``hints'' (indices inline, indices in output) to be slightly more abstract; 
and purely symbolic input representations (AMR graphs) as most abstract. 
Other I/O representations are ranked based on the output representation: BIO tags are more abstract than indices (they require rules for matching); 
and both are more complex than concrete settings, since outputting a matching string is easier than mapping to an arbitrary symbol. 
The list of experiments is in \tabref{representations}, and our results in \figref{abstraction}. 

For non-CoT approaches, we observed noticeable improvements in low-abstraction scenarios (indices inline and indices in output; concrete). 
As it rose, LLMs scored worse, though remaining above random. 
The results are significant under a Welch's $t$-test on the prediction arrays. 
When the input text is unaltered, line numbers or BIO tags make no difference in predictions with CoT ($p \approx 0.77$; large $p$-values imply the distributions have identical expected values), but are noticeably worse without ($p < 0.05$). 
Another $t$-test shows that $t < -0.86$ and $p \approx 0.39 $ for inline indices when compared with its non-CoT version. 
Hence we reject the null hypothesis that the distributions are distinct. 
Since the performance in non-CoT was better on average (67\% vs 70\% \Fw), it is possible that CoT harms performance on outlier points. 
We compared CoT in other scenarios, and also observed large $p$ values when comparing with only indices in the output ($<0.98$); and when comparing the latter with BIO tags ($p < 0.77$). 
This suggests that CoT is effective in highly abstract scenarios, such as requesting BIO tags, but detrimental otherwise. 

\section{Discussion}\label{sec:discussion}

The LLMs scored highly in AM and APE, to the point of beating or almost matching the existing SOTA models. 
This is not sufficient to claim that the models are able to perform argumentative reasoning. %
When we altered the I/O representation with conceptually minor changes--such as adding line numbers or requesting a BIO set as opposed to integers--the LLMs had noticeably different performances. 
An explanation for the relatively low scores in APE is that the LLMs failed to generalize due to the length of the task; which causes problems in transformer-based models \citep{anil2022exploring}. 

CoT prompts had \emph{on average} higher scores than their non-CoT counterparts. 
They also yielded better scores in ill-posed (overly abstract) scenarios. 
However, our analysis showed that the output distributions for all CoT approaches were rather similar. 
This is perhaps indicative that CoT allows the models to return the same output regardless of representation. 
This appears to be due to the templatized nature of CoT, i.e., ``Let's read line-by-line and solve step-by-step'', and the specific steps needed to generate the output regardless of the I/O representation. 
This itself is not indicative of reasoning. 

Finally, there is a clear peak at four exemplars with respect to the model's downstream performance. 
This suggests that, assuming that more exemplars imply better information about the task, the LLMs are not accurately performing inference from the data provided. 
This exemplar effect did not extend to CoT settings, %
though we do not discard the possibility that models with longer token limitations could also show this trend. 
Overall, we pose that the models are unable to reason in an argumentative setting, but their scores give an excellent appearance of being able to do so. 

\section{Limitations}\label{sec:limitations}
Our analysis has three main limitations. 
In terms of reasoning evaluation, it could be argued that our results are not complete in terms of evaluating argumentative reasoning capabilities. 
We agree: \emph{recognizing} an argument is not the same as \emph{deciding} its quality. 
However, without the ability of the model to show that it is able to recognize arguments and identify relations between them, any potentially generated argument or result evaluating the model's performance in these tasks is untrustworthy. 

We factored out, to an extent, potential data contamination, which is known to impact downstream model performance \citep{PlagiariseLee,dewynter2023evaluation}, by tasking the model to recognize arguments from the passage. 
However, we are unable to guarantee that the models have not been trained with this data, and therefore have at least some bias towards these results. 
Finally, we only evaluated two models, so our results may not extend to other LLMs. 
Likewise, we did not fine-tune the models, and opted instead to treating them as generalists performing in-context learning, in line with their contemporary usage. 

\section{Conclusion}\label{sec:conclusion}

We evaluated the argumentative reasoning capabilities of GPT-3 and GPT-4, by measuring whether they could recognize arguments from a passage--the first step on performing such reasoning. 
The LLMs score well in AM and APE, beating or nearly-matching the SOTA. 
However, statistical analysis on the LLMs' predictions when subject to small, yet still human-readable, alterations in the I/O representations showed that the LLMs were extremely sensitive to the abstraction level and the number of exemplars. 
Hence, we concluded that they were not reasoning over the arguments seen. 

However, symbolic prompting strategies (e.g. reducing the abstraction level of the prompt by adding line numbers) allowed the LLMs to score well and even beat CoT. 
We were also unable to conclude that CoT helped argumentative reasoning in LLMs, but did observe more robust results due to its templatized nature. 
Our analysis implies that it helps mitigate issues stemming from overly abstract or ill-conditioned problems. 

As mentioned in \secref{limitations}, 
we did not evaluate the LLMs' ability to judge an argument's strength, or to provide reasonable rebuttals. 
Moreover, due to token length limitations, we were unable to evaluate AMR, the most abstract setting we tested, with CoT. 
We believe that this evaluation could provide valuable insights on to what extent these models are able to discern abstract input representations. 
Further work could explore these issues. 
Overall, our work suggests that LLM usage in areas like data labelling and paper reviewing must be exercised with care and good judgement.

\section*{Acknowledgements}

The authors wish to thank Liying Cheng for answering questions about the RRv2 dataset. 

\bibliographystyle{vancouver}
\DeclareRobustCommand{\DE}[3]{#2}
\bibliography{biblio}

\appendix

\section{Full Prompts}
In this section we show samples of our AM and APE prompts.\footnote{Full prompts in \url{https://github.com/adewynter/argumentation}}. 
For AM we include our concrete (\figref{amconcrete}), symbolic (BIO and indices inline; \figstworef{amsymbolicbio}{amsymbolicindices}), and one CoT (BIO; \figref{amsymbolicbiocot}) prompts. 
For APE we include the concrete (\figref{apeconcrete}) and one symbolic (indices as output; \figref{apesymbolicindices}) prompts. 
Due to space, we have omitted sample inputs and part of the exemplars.

\begin{figure}
\centering
\begin{promptbox}
\begin{lstlisting}
Here is a |review| and a |response|. Return all pairs
of argument and rebuttal from the text.
Arguments must come from the |review|.
Rebuttals must come from the |response|.

Example:
|start of review|
{EXEMPLAR REVIEW 1 GOES HERE}
 |end of review|
|start of response|
{EXEMPLAR RESPONSE 1 GOES HERE}
|end of response|
Answer:
|start of answer|
- argument 0 (5 6) : 1 2 3 4 5 6 7 8 9 10 11 12 13
|end of answer|

Passages begin:
|start of review|
{REVIEW}
 |end of review|
|start of response|
{RESPONSE}
|end of response|
Answer:
|start of answer|
\end{lstlisting}
\end{promptbox}
\caption{APE symbolic prompt with one exemplar (omitted). 
There is only one match as a possible solution (lines $5,6$ in the review; lines $1-13$ in the response). 
In zero-shot scenarios we specify the desired output representation.}
\label{fig:apesymbolicindices}
\end{figure}

\begin{figure}
    \centering
\begin{promptbox}
\begin{lstlisting}
Here is a passage.
Return every line from the passage that is part of an
argument, and ONLY these lines. Mark the beginning of
every argument with |START|.

For example:
|passage start|
- The author studies the quantization strategy of
CNNs in terms of Pareto Efficiency.
- Through a series of experiments with three 
standard CNN models (ResNet, VGG11, MobileNetV2),
the authors demonstrated that lower precision value
can be better than high precision values in term of
Pareto efficiency under the iso-model size scenario.
- They also study cases with and without depth-wise
convolution, and propose a new quantization method,
DualPrecision.
...
- Comment:
- I am not at all familiar with quantization methods, 
therefore no knowledge of relevant related works.
- If, however, the authors did a thorough job of
surveying related works and chose sensible baselines,
I think the experiments demonstrate the usefulness
of the new DualPrecision technique.
|passage end|
Response:
- |START| I am not at all familiar with quantization
methods, therefore no knowledge of relevant related
works.
- If, however, the authors did a thorough job of
surveying related works and chose sensible baselines,
I think the experiments demonstrate the usefulness of
the new DualPrecision technique.

Passage:
|passage start|
{PASSAGE}
|passage end|
Response:
- \end{lstlisting}
\end{promptbox}
\caption{AM concrete prompt with one exemplar. 
We replace the \{PASSAGE\} field with the input passage. 
The output text has a special token (``\texttt{|START|}'') for parsing.}
\label{fig:amconcrete}
\end{figure}

\begin{figure}
    \centering
\begin{promptbox}
\begin{lstlisting}
Here is a passage.
For every line in the passage:
- Mark the beginning of every argument with B.
- Mark the inside of every argument with I.
- Mark everything else with O.

For example:
|passage start|
{EXEMPLAR GOES HERE}
|passage end|
Response:
Let's think step-by-step and read it line-by-line.
We read: ``The author studies the quantization
strategy of CNNs in terms of Pareto Efficiency.'': 
This sentence is not related to an argument, so we
mark it as ``O''.
...
Next we read: ``If, however, the authors did a
thorough job of surveying related works and chose
sensible baselines, I think the experiments
demonstrate the usefulness of the new DualPrecision
technique.'': This follows the previous sentence. So
we are still reading an argument. We mark it as 
``I''.
So the answer should be:
 - O
 - O
 - O
 - O
 - O
 - B
 - I

Passage:
|passage start|
{PASSAGE}
|passage end|
Response:
Let's think step-by-step and read it line-by-line.
We read: ``
 - \end{lstlisting}
\end{promptbox}
        \caption{AM symbolic (indices) prompt with one exemplar and CoT. Refer to \figref{amconcrete} for a longer version of the exemplar. 
        This prompt performs step-by-step reasoning on AM by following a templatized generation and specific labelling logic. 
        We replace the \{PASSAGE\} field with the desired input, and begin parsing when we observe either a ``So the answer should be:'' string, or the last elements of each sentence in case the former is not found.}
    \label{fig:amsymbolicbiocot}
\end{figure}

\begin{figure}
    \centering
    \begin{promptbox}
    \begin{lstlisting}
Here is a passage.
For every line in the passage:
- Mark the beginning of every argument with B.
- Mark the inside of every argument with I.
- Mark everything else with O.

For example:
|passage start|
|passage start|
- The author studies the quantization strategy of
CNNs in terms of Pareto Efficiency.
- Through a series of experiments with three 
standard CNN models (ResNet, VGG11, MobileNetV2),
the authors demonstrated that lower precision value
can be better than high precision values in term of
Pareto efficiency under the iso-model size scenario.
- They also study cases with and without depth-wise
convolution, and propose a new quantization method,
DualPrecision.
- DualPrecision empirically outperformed 8-bit
quantization and flexible quantization methods on
ImageNet.
- Comment:
- I am not at all familiar with quantization methods, 
therefore no knowledge of relevant related works.
- If, however, the authors did a thorough job of
surveying related works and chose sensible baselines,
I think the experiments demonstrate the usefulness
of the new DualPrecision technique.
|passage end|
Response:
 - O
 - O
 - O
 - O
 - O
 - B
 - I

Passage:
|passage start|
{PASSAGE}
|passage end|
Response:
 - \end{lstlisting}
    \end{promptbox}
    \caption{AM symbolic (BIO) prompt with one exemplar and no CoT. 
    Refer to \figref{amconcrete} for a longer version of the exemplar. 
    We replace the \{PASSAGE\} field with the desired input, and begin parsing when we observe a ``Response:'' string, or one of the BIO tags.}
    \label{fig:amsymbolicbio}
\end{figure}

\begin{figure}
    \centering
\begin{promptbox}
\begin{lstlisting}
Here is a |review| and a |response|.
Return all pairs of argument and rebuttal from the
text.
Arguments must come from the |review|.
Rebuttals must come from the |response|.

Example:
|start of review|
{EXEMPLAR REVIEW 1 GOES HERE}
 |end of review|
|start of response|
{EXEMPLAR RESPONSE 1 GOES HERE}
|end of response|
Answer:
|start of answer|
- argument 0
 - I am not at all familiar with quantization methods
therefore no knowledge of relevant related works.
  - If, however, the authors did a thorough job of
 surveying related works and chose sensible baselines
 I think the experiments demonstrate the usefulness
 of the new DualPrecision technique.
- rebuttal 0
 - We have updated our manuscript so that it states
 our contributions explicitly.
  - While our proposed DualPrecision is simple yet
  effective, we would like to emphasize that it is
  only one of our contributions.
  - For the reviewer's convenience, we re-iterate our
  main contributions below:
  - Overall, we systematically analyze the model size
  and accuracy trade-offs considering both weight
  precision values and the number of channels for
  various modern networks architectures (variants of
  ResNet, VGG, and MobileNet) and datasets (CIFAR
  and ImageNet) and have the following non-trivial
  and novel contributions:
...
|end of answer|

Passages begin:
|start of review|
{REVIEW}
|end of review|
|start of response|
{RESPONSE}
|end of response|
Answer:
|start of answer|
\end{lstlisting}
\end{promptbox}
        \caption{APE concrete prompt (non-zero shot, but the exemplars have been omitted for brevity). 
        The response is given in Markdown format, with the indentation denoting the interior parts of an argument. 
        We also use an indexing system (``argument 0'', ``rebuttal 0'') for matching. 
        }
    \label{fig:apeconcrete}
\end{figure}

\begin{figure}
    \centering
\begin{promptbox}
\begin{lstlisting}
Here is a passage.
For every line in the passage:
- Group and return the index of lines corresponding
to an argument.
- Put the index of the beginning of the argument in
parenthesis, and then list below the remaining
indices, if any.
- Do not return anything else.

For example:
|passage start|
- 0) The author studies the quantization strategy of
CNNs in terms of Pareto Efficiency.
- 1) Through a series of experiments with three
standard CNN models (ResNet, VGG11, MobileNetV2),
the authors demonstrated that lower precision value
can be better than high precision values in term of
Pareto efficiency under the iso-model size scenario.
- 2) They also study cases with and without depth- 
wise convolution, and propose a new quantization
method, DualPrecision.
- 3) DualPrecision empirically outperformed 8-bit
quantization and flexible quantization methods on
ImageNet.
- 4) Comment:
- 5) I am not at all familiar with quantization methods,
therefore no knowledge of relevant related works.
- 6) If, however, the authors did a thorough job of
surveying related works and chose sensible baselines,
I think the experiments demonstrate the usefulness of
the new DualPrecision technique.
|passage end|
Response:
 - (5)
 - 6

Passage:
|passage start|
{PASSAGE}
|passage end|
Response:
 - \end{lstlisting}
\end{promptbox}
        \caption{AM symbolic (indices inline, and indices as output) prompt with one exemplar and no CoT. 
        This prompt improves the input representation by including indices as part of the input in a predefined format (parentheses). 
        We replace the \{PASSAGE\} field with the desired input, and begin parsing when we observe a ``Response:'' token. To tell labels apart, we have requested the beginning of the argument to be enclosed in parentheses.}
    \label{fig:amsymbolicindices}
\end{figure}

\end{document}